\documentclass[10pt, conference, compsocconf]{IEEEtran}
\usepackage{cite}
\usepackage{amsmath}
\usepackage{balance}
\usepackage{amssymb}
\usepackage[utf8]{inputenc}
\usepackage{color, soul}
\usepackage{url}
\usepackage{textcomp}
\usepackage{tabularx}
\usepackage{graphicx}
\usepackage[caption=false]{subfig}
\usepackage{wrapfig}

\def\BibTeX{{\rm B\kern-.05em{\sc i\kern-.025em b}\kern-.08em
    T\kern-.1667em\lower.7ex\hbox{E}\kern-.125emX}}
\DeclareRobustCommand*{\IEEEauthorrefmark}[1]{%
  \raisebox{0pt}[0pt][0pt]{\textsuperscript{\footnotesize #1}}%
}

\ifCLASSINFOpdf
\else
\fi
\begin{document}
\IEEEoverridecommandlockouts
%
% paper title
% can use linebreaks \\ within to get better formatting as desired
\title{Machine Learning Algorithms for Classification of Microcirculation Images from Septic and Non-Septic Patients}

% author names and affiliations
% use a multiple column layout for up to two different
% affiliations

\author{
    \IEEEauthorblockN{
        Perikumar Javia\IEEEauthorrefmark{1}, 
        Aman Rana\IEEEauthorrefmark{1},
        Nathan Shapiro\IEEEauthorrefmark{2}, and
        Pratik Shah\IEEEauthorrefmark{1}$^{\dagger}$\thanks{$^{\dagger}$Corresponding author: pratiks@media.mit.edu}}
        
    \IEEEauthorblockA{
        \begin{tabular}{cc}
            \begin{tabular}{@{}c@{}}\\
                \IEEEauthorrefmark{1}
                    MIT Media Lab\\
                    Massachusetts Institute of Technology\\
                    Cambridge, MA, USA\\
                    \textit{\{pjavia, arana, pratiks\}@media.mit.edu}
            \end{tabular} & \begin{tabular}{@{}c@{}}\\
                \IEEEauthorrefmark{2}
                	Emergency Medicine\\
                    Beth Israel Deaconess Medical Center\\
                    Boston, MA, USA\\
                    \textit{nshapiro@bidmc.harvard.edu}
            \end{tabular}
        \end{tabular}
    }
\thanks{Published at 2018 17th IEEE International Conference on Machine Learning and Applications}
\thanks{© 2018 IEEE.  Personal use of this material is permitted.  Permission from IEEE must be obtained for all other uses, in any current or future media, including reprinting/republishing this material for advertising or promotional purposes, creating new collective works, for resale or redistribution to servers or lists, or reuse of any copyrighted component of this work in other works.}
\thanks{DOI: 10.1109/ICMLA.2018.00097}
}

% conference papers do not typically use \thanks and this command
% is locked out in conference mode. If really needed, such as for
% the acknowledgment of grants, issue a \IEEEoverridecommandlockouts
% after \documentclass

% for over three affiliations, or if they all won't fit within the width
% of the page, use this alternative format:
% 
%\author{
%    \IEEEauthorblockN{Aman Rana$^{1}$} \and
%    \IEEEauthorblockN{Gregory Yauney$^{1}$} \and
%    \IEEEauthorblockN{Alarice Lowe$^{2}$} \and
%    \IEEEauthorblockN{Pratik Shah$^{1}$}
%
%    \IEEEauthorblockA{
%    \begin{tabular}
%        $^{2}$Massachusetts Institute of Technology,
%        Cambridge, MA, USA\\
%        \{arana,gyauney,pratiks\}@media.mit.edu
%    } \and
%    \IEEEauthorblockA{$^{2}$Brigham and Women's Hospital\\
%        Harvard Medical School,
%        Boston, MA, USA\\
%        alowe@bwh.harvard.edu
%    }
%}

% make the title area
\maketitle

\begin{abstract}
Sepsis is a life-threatening disease and one of the major causes of death in hospitals. Imaging of microcirculatory dysfunction is a promising approach for automated diagnosis of sepsis. We report a machine learning classifier capable of distinguishing non-septic and septic images from dark field microcirculation videos of patients. The classifier achieves an accuracy of 89.45\%. The area under the receiver operating characteristics of the classifier was 0.92, the precision  was 0.92 and the recall was 0.84. Codes representing the learned feature space of trained classifier were visualized using t-SNE embedding and were separable and distinguished between images from critically ill and non-septic patients. Using an unsupervised convolutional autoencoder, independent of the clinical diagnosis, we also report clustering of learned features from a compressed representation associated with healthy images and those with microcirculatory dysfunction. The feature space used by our trained classifier to distinguish between images from septic and non-septic patients has potential diagnostic application.
\end{abstract}


% For peer review papers, you can put extra information on the cover
% page as needed:
% \ifCLASSOPTIONpeerreview
% \begin{center} \bfseries EDICS Category: 3-BBND \end{center}
% \fi
%
% For peerreview papers, this IEEEtran command inserts a page break and
% creates the second title. It will be ignored for other modes.
\IEEEpeerreviewmaketitle

\section{Introduction}
\begin{figure*}[t]
\centerline{\includegraphics[scale=0.5]{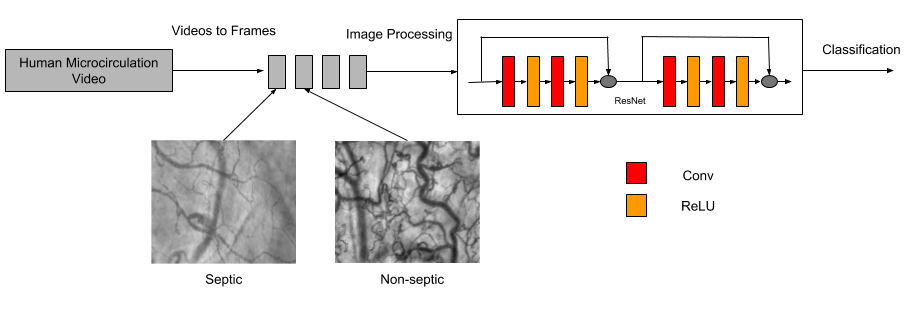}}
\caption{Frames were sampled from microcirculation videos, resized to 224 x 224 x 3 pixels and fed into CNN to distinguish images from non-septic and septic patients. A truncated version of ResNet18 was used as architecture.}
\label{fig1}
\end{figure*}

Sepsis is a life-threatening disease where the host response to an infection leads to inflammation that may result in multiple organ failure and death of the patient \cite{b1}. Sepsis is associated with very high mortality rates that make early detection crucial for treatment. Microcirculation is the flow of blood in the smallest elements of the cardiovascular system, the capillary network, where the exchange of oxygen takes place and is pivotal in the maintenance of homeostasis \cite{b2}. Dysfunctions in microcirculation system, such as changes in flow of blood in the capillary network, constriction of capillary vessel etc. manifest early in the pathophysiology of sepsis. Previous studies demonstrate as association between microcirculatory derangements and sepsis severity and outcomes \cite{b3} \cite{b14}.

Improvements in microcirculation function after early resuscitation were associated with reduction in subsequent organ failure \cite{b3} \cite{b16} \cite{b13}. This evaluation of microcirculation function holds potential for treating patients with sepsis. Invasive methods such as serum lactate physiology are often used as indirect measures for the evaluation of the microcirculatory function \cite{b3}. Direct clinical evaluation of microcirculatory dysfunction using dark field imaging, a relatively fast and non-invasive approach, has been investigated by researchers for diagnosis and prognosis of sepsis. Measurements relating to capillary density and flow in side-stream dark field (SDF) videos have been used to evaluate microcirculatory dysfunction either by human raters, simple software systems or computer vision methods \cite{b3} \cite{b10} \cite{b11} \cite{b17}.

Machine learning has been used successfully to automate diverse tasks such as recognition of objects and semantic segmentation in self-driving cars. Deep learning architectures such as convolutional neural networks (CNNs) can learn to segment distinguishing features in images with minimal human supervision \cite{b4}. CNNs are comprises of layers of stacked neurons where input features are transformed into higher dimensional representation using convolution.  The convolution is often wrapped with non-linear activation function. Max pooling is usually used for downsampling and the final compressed high dimensional features are fed to softmax classifier for classification. CNNs have achieved high accuracy in skin cancer detection from RGB images, lung nodule classification using CT images and segmentation of periodontal diseases from oral images \cite{b5} \cite{b6} \cite{b7} \cite{b8}. Previous work has used deep learning and computer vision to investigate microcirculatory dysfunction in dark field imaging videos of pigs and other animals \cite{b9} \cite{b12}. To our knowledge no previous study has successfully used deep learning to detect microcirculation dysfunction from critically ill human subjects.

In this work, we implement a CNN to analyze microcirculatory dysfunction captured by dark field imaging in human patients and distinguish between septic and non-septic images with high accuracies. We also investigate outputs from the last convolutional layer in our architecture, and using a t-SNE embedding show that the representation learned by the classifier successfully differentiates the frames. Additionally, an unsupervised learning approach, independent of clinical labels, was used to investigate the feature space of the microcirculation image validation dataset and showed clustering of images from non-septic and septic patients. To our knowledge, this is the first study that successfully classifies human microcirculation image data using a deep neural network. We reason the salient feature space used by our trained classifier may have diagnostic applications.

\section{Related Work}

In previous studies, microcirculation dysfunction was experimentally induced in animal models with drugs and other agents followed by SDF imaging \cite{b12}. SDF imaging is a non-invasive imaging modality and has been used to track changes in the microcirculation on mucosal surfaces. SDF uses green polarized light with wavelength of 550 nm which is absorbed by hemoglobin and makes red blood cells visible \cite{b22}. The distinguishing parameters and measurements associated to microcirculation include MFI (Microcirculation Flow Index), PVD (Perfused Vessel Density), TVD (Total Vessel Density) and PPV (Portion of Perfused Vessels) \cite{b3}. These measurements require identification of vessels in microcirculation video frames. Several software systems have been developed to analyze the microcirculation images and videos for these measurements, but often fail to provide desired results \cite{b18}. Marking of hand engineered features on each frame of the video is required at some stage of analyses and is time-consuming and not accurate \cite{b17}.

    Computer vision techniques have previously been used to automate  microcirculation parameters and associated measurements to detect dysfunction. Liu et. al report a framework to estimate the flow rate statistics from microcirculation SDF imaging after the video is stabilized, enhanced and micro-vessels are extracted from each frame \cite{b9}.  Similarly, Bezemer et. al report an automated assessment of the TVD measurement using contrast score thresholding to validate the vessel detection on microcirculation videos collected from patients \cite{b11}. Demir et. al automate the estimation of functional capillary density (FCD) by preprocessing multi-thresholding segmentation on four blocks of five consecutive frames; followed by morphological operations on the mean image of each block and finally region growth models to extract the vessels to calculate the FCD \cite{b10}. Graph-based approaches have also been leveraged to extract features from flow and capillary structure to assess microcirculation dysfunction caused due to sepsis in mice \cite{b19}. Mcllroy et. al reported use of machine learning to classify microcirculation videos from animals into discrete intensities of drug induced microcirculation inflammation. They report no significant differences in prediction accuracy when comparing the performance of training on videos vs. frames \cite{b12}.
    
\section{Methods}

\subsection{Data Acquisition and Processing}

Deidentfied microcirculation videos were acquired from BIDMC (Beth Israel Deaconess Medical Centre, Boston, MA) collected using protocol 2008P-000089. De-identified videos were transferred to Massachusetts Institute of Technology and analyzed according to Committee on the Use of Humans as Experimental Subjects protocol 1705964591. Videos were captured using a CytoCam Incidence Dark Field imaging camera system (Braedius Medical B.V., The Netherlands). Videos were acquired from consenting patients in the Emergency Department of Intensive care unit (ICU) who were diagnosed with septic shock (low blood pressure from sepsis). Non-septic consenting patients were also imaged as controls. The videos were acquired by placing the imaging device under the tongue of the patients and were recorded at two different intervals for septic patients i.e. at zero hours and 12 hours after study enrollment. Control videos were taken only at zero hours. The diagnosis of patients as non-septic or septic following clinical evaluation was considered as the ground truth for training and validation of machine learning classifiers but not for unsupervised learning using autoencoders. 

\begin{table}[htbp]
\caption{Data Splits}
\begin{center}
\begin{tabular}{|c|c|c|}
\hline
& \textbf{Number of} &\textbf{Number of} \\ 
\textbf{Label} & \textbf{training frames} & \textbf{validation frames} \\
\hline
Septic & 8191 & 2410  \\
Non-septic & 14026 & 2613 \\
\hline
%\hline
\end{tabular}
\label{tab1}
\end{center}
\end{table}

Raw videos were reviewed for quality control and videos frames that did not meet expected standards of illumination, focus, duration, content, stability and pressure were discarded \cite{b18}. In total 133 videos acquired from 53 patients were selected for machine learning. Videos were stabilized and cropped to remove unnecessary image background. Frames were extracted from the videos and split into 80\% for training and 20\% for validation (Table 1).  All the frames from a single patient were used either for training or the validation dataset to prevent over fitting on patient-specific signatures.

\subsection{Classifier}

ResNet18 was modified by removing layers, resulting in a 10-layer architecture used in this study, while retaining the ResNet structure \cite{b15}. Due to variability in size of videos frames were cropped and were resized to 224 x 224 x 3 pixels to create a uniform dataset. A batch size of four was used during training and Adam optimizer was used as an optimizer \cite{b20}. Cross Entropy was used as the loss metric. The training process is shown in Fig. 1. NVIDIA GeForce GTX Titan X GPU was utilized to speed up the training time and the code was written using PyTorch library.

\subsection{Visualization of CNN codes}

t-SNE was used to visualize the convolutional feature vectors for the validation dataset in two dimension \cite{b23}. After the architecture was trained layers were frozen and the frames in the validation set were fed to the architecture. The 128-dimensional output of the last convolutional layer (convolutional feature vector) was used for t-SNE embedding and visualization (Fig. 2).

\begin{figure}[h]
\centerline{\includegraphics[scale=0.4]{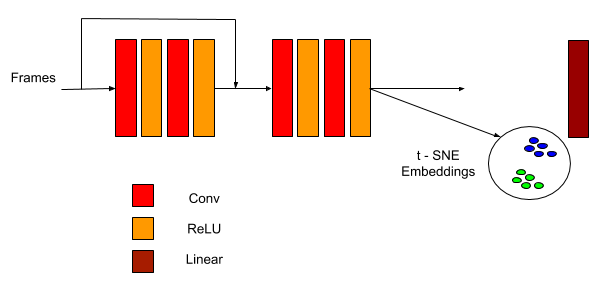}}
\caption{The last convolutional layer used for visualization of t-SNE embedding.}
\label{fig2}
\end{figure}

\subsection{Autoencoders}

For investigating the feature space of the extracted frames, an unsupervised convolutional autoencoder was trained with reconstruction loss (Fig. 3). The autoencoder consisted of convolutional layers in the encoder followed by ReLU and BatchNorm \cite{b24}. The encoder and decoder comprised of three convolutional layers each. No skip connections were used. The autoencoder was trained using Adam optimizer and mean squared error as the loss metric. The training and test dataset remained unchanged but clinical labels were not used. A few selected images from the validation dataset were randomly sampled after every epoch to monitor the similarity between input and output frames. Once autoencoder was trained, the feature vector (1152 dimensional bottleneck layer) was extracted and k-means algorithm was used for clustering and visualization \cite{b21}. For designing bottleneck layer the number of neurons were kept as few as possible in order to avoid identity mapping and still achieve better reconstruction loss.

\begin{figure}[h]
\centerline{\includegraphics[scale=0.4]{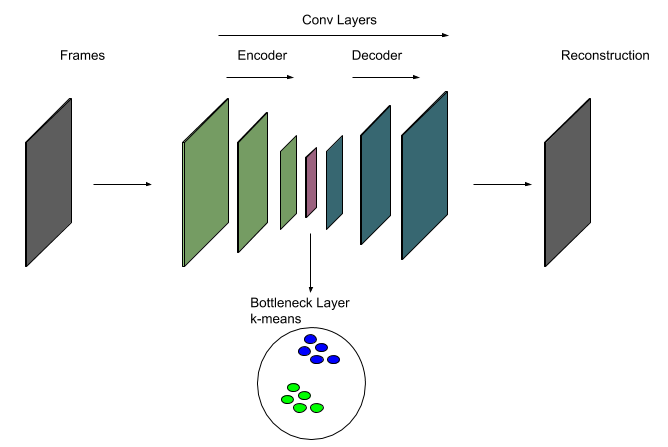}}
\caption{Feature vector from the trained autoencoder used for k-means clustering and visualization.}
\label{fig3}
\end{figure}

\section{Results}

We report a machine learning classifier that successfully distinguishes between non-septic patients and septic frames extracted from SDF microcirculation videos of septic and non-septic patients. The classifier achieves an accuracy of 89.45\% on a validation dataset comprising of 5,023 frames. The area under the receiver operating characteristic curve (AUROC) was 0.92 (Fig. 4). Supplementary table \ref{table: supp_tab_1} shows mapping of correctly classified frames to individual patients in the validation dataset. Although the total number of septic and non-septic patients were not equal, the total number of frames were balanced as shown in Table \ref{tab1}. 

\begin{figure}[h]
\centerline{\includegraphics[scale=0.3]{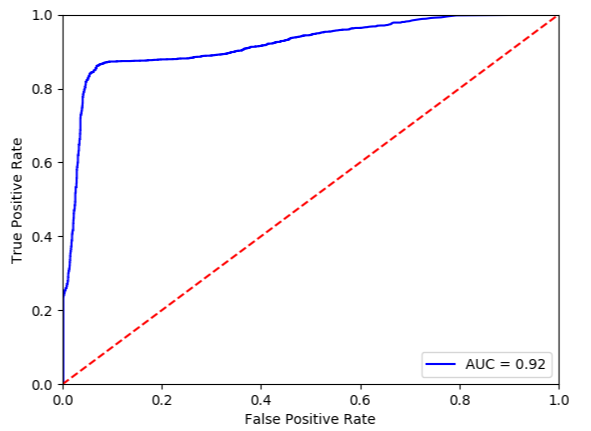}}
\caption{Receiver operating characteristic curve.}
\label{fig4}
\end{figure}

\newpage
\begin{figure}[ht]
\centerline{\includegraphics[scale=0.4]{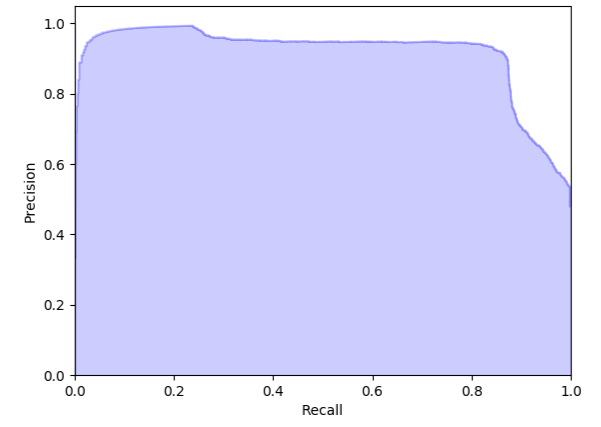}}
\caption{Precision-recall curve for the validation data.}
\end{figure}

The precision and recall were 0.92 and 0.84 respectively (Fig. 5). Figure 6 shows the t-SNE embedding of the convolutional features on the validation set, with  visually separable clusters. Thus, showing that CNN does indeed separate images from septic and non-septic patients into distinct clusters for classification. The results of autoencoder were measured based on reconstruction loss in terms of mean squared error, with a value of 0.005362; thus, showing that the sampled decoded images visually matched the input images. The accuracy on validation set using k-means clustering was 73.32\%.

\section{Discussion}

The AUROC of the CNN classifier is 0.92 (Fig. 4), which indicates that the accuracy of the classifier is above chance. This also shows that there are distinct learnable representations, independent of temporal information, which can differentiate between frames from non-septic and septic patients. The classifier achieves a precision value of 0.92, indicating that 92\% of its predictions were accurately classified. The classifier was able to recall frames from septic and non-septic patients correctly 84\% of the time. Our patient-level data in supplementary Table \ref{table: supp_tab_1} shows that seven patients in the validation dataset had more 90\% of their images classified accurately, with 100\% accuracy for four individuals. Images from two patients were classified with low accuracies. Variability in image capture and background have been reported as potential confounding factors precluding clinical evaluation. 
Inherent heterogeneity in patient anatomical features could also play a role in reducing classification accuracies. Images from a single patient were used either for training or in the validation dataset as one potential approach to prevent over fitting on patient-specific signatures.

Stuart McIlroy et. al report an accuracy of 83\% when detecting inflammation in microcirculation videos of animals; and 80\% accuracy when classifying the degree of inflammation \cite{b12}. They train a CNN model on microcirculation videos of animals with drug induced inflammation. We use microcirculation videos from human subjects where the inflammation was induced as the body’s natural response to infection. We
\begin{figure}[ht]
\centerline{\includegraphics[scale=0.45]{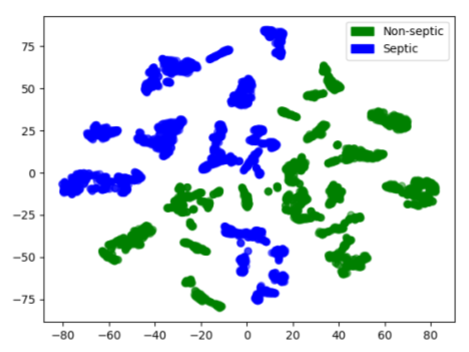}}
\caption{Visualization of t-SNE embedding using the last convolutional layer of the trained CNN model.}
\label{fig6}
\end{figure}

\noindent
report an accuracy of 89.45\% when detecting presence of sepsis in the images and also show t-SNE embedding of the features learned by the convolutional layers. We further extend our work by showing that an unsupervised autoencoder can distinguish between images from septic and non-septic patients without using clinical labels. Thus, our work is significantly different compared to this study and novel.

Previous studies use temporal information to detect blood flow in the capillaries \cite{b9} \cite{b10} \cite{b11} \cite{b12}. We on the other hand use static frames extracted from the microcirculation videos for classification. The first step for the previously described studies in the field required detection of vessels in the videos, followed by calculation of measurements such as MFI, PVD, TVD etc, which were used to evaluate dysfunction in microcirculation. Hand engineered features, image segmentation and computer vision algorithms have also been used to extract vessels in the videos \cite{b3} \cite{b10} \cite{b11} \cite{b17}. While we, use a CNN to automatically learn features in the raw frames to classify septic vs. non-septic patients. We hypothesize that our neural network may be learning features, that are independent of temporal information, such as variation in numbers, length and density of vessels between images from septic and non-septic patients. Previous work by Stuart McIlroy et. al has also alluded about the dispensbility of temporal learning for successfully classification in animals. The overlap, if any, between diagnostic features used by human experts and machine learning algorithms is an active area of investigation in our research group.

Visualization and examination of the convolutional features learned by the CNN has been shown (Fig. 6). The cluster separation indicate that the classifier had learned to effectively distinguish between images of septic and non-septic patients. The slight differences in the clustering could be attributed to variability in illumination, stability, contrast between images. The features learned by bottleneck layer of the unsupervised autoencoder were extracted using reconstruction loss, and thereby are not influenced by the sepsis labels provided by clinicians. The accuracy of 73.32\% using the k-means algorithm on these features show that the images indeed had general features that can be leveraged or refined to distinguish between septic and non-septic frames.

\section{Conclusion}

We report the first CNN classifier that successfully classifies human SDF microcirculation images with high accuracy. Representation and visualization of the learned embeddings of the trained classifier supports the conclusion that it may use unique features to distinguish between immages from septic or non-septic patients. Additionally, we use an unsupervised autoencoder to show that indeed there may be differential features in human microcirculation. We propose that the salient feature space used by our trained classifier has diagnostic applications for evaluating microcirculation dysfunction in humans.

\section*{Acknowledgment}
Gregory Yauney and Michael Massey for technical assistance

\setcounter{table}{0}
\renewcommand{\thetable}{S\arabic{table}}

\balance

\section*{Supplementary Material}
\begin{table}[htbp]
\caption{Patient Level accuracy distribution}
\begin{center}
\begin{tabular}{|c|c|c|c|c|c|}
\hline
 & \textbf{Total} & \textbf{Correct}& \textbf{Incorrect}&\textbf{Percent} & \textbf{Clinical} \\

\textbf{Patient} & \textbf{frames} & \textbf{pred.} & \textbf{pred.} & \textbf{accuracy} & \textbf{label} \\

\hline
P1 & 377 & 377 & 0 & 100 \% & Septic \\
P2 & 961 & 957 & 4 & 99.58 \% & Septic \\
P3 & 1072 & 714 & 358 & 66.60 \% & Septic \\
P4 & 958 & 954 & 4 & 99.58 \% & Non-septic \\
P5 & 135 & 135 & 0 & 100 \% & Non-septic \\
P6 & 519 & 463 & 56 & 89.21 \% & Non-septic \\
P7 & 194 & 86 & 108 & 44.32 \% & Non-septic \\
P8 & 300 & 300 & 0 & 100 \% & Non-septic \\
P9 & 507 & 507 & 0 & 100 \% & Non-septic \\
\hline
\end{tabular}
\label{table: supp_tab_1}
\end{center}
\end{table}

\end{document}